\documentclass[runningheads]{llncs}

\usepackage{multirow}
\usepackage{graphicx}
\usepackage{array}
\usepackage{amssymb}
\usepackage{paralist}
\newcolumntype{P}[1]{>{\centering\arraybackslash}p{#1}}
\usepackage{booktabs}
\usepackage[normalem]{ulem}
\usepackage[rgb]{xcolor}

\newcommand\colorsquare[4]{\textcolor[HTML]{#1}{\rule{#3}{#4} #2}}

\usepackage{todonotes}
\usepackage{hyperref}
\usepackage[capitalise]{cleveref}
\usepackage{censor}

\usepackage{url}

\usepackage{todonotes}

\usepackage{orcidlink}

\begin{document}

\title{
QAGCN: Answering Multi-Relation Questions via Single-Step Implicit Reasoning over Knowledge Graphs
}

\titlerunning{QAGCN: Answering Multi-Relation Questions over Knowledge Graphs}

\author{
Ruijie Wang\inst{1,2}\orcidlink{0000-0002-0581-6709} \and
Luca Rossetto\inst{1}\orcidlink{0000-0002-5389-9465} \and
Michael Cochez\inst{3,4}\orcidlink{0000-0001-5726-4638} \and \\
Abraham Bernstein\inst{1}\orcidlink{0000-0002-0128-4602}
}
\authorrunning{R. Wang et al.}
%
\institute{
Department of Informatics, University of Zurich, Zurich, Switzerland
\and University Research Priority Program ``Dynamics of Healthy Aging'',\\ University of Zurich, Zurich, Switzerland
\\
\email{\{ruijie, rossetto, bernstein\}@ifi.uzh.ch}
\and
Vrije Universiteit Amsterdam, Amsterdam, The Netherlands
\and
Discovery Lab, Elsevier, Amsterdam, The Netherlands
\\
\email{m.cochez@vu.nl}
}

\maketitle

\begin{abstract}
Multi-relation question answering (QA) is a challenging task, where given questions usually require long reasoning chains in KGs that consist of multiple relations.
Recently, methods with explicit multi-step reasoning over KGs have been prominently used in this task and have demonstrated promising performance.
Examples include methods that perform stepwise label propagation through KG triples and methods that navigate over KG triples based on reinforcement learning.
A main weakness of these methods is that their reasoning mechanisms are usually complex and difficult to implement or train.
In this paper, we argue that multi-relation QA can be achieved via end-to-end single-step implicit reasoning, which is simpler, more efficient, and easier to adopt.
We propose QAGCN --- a Question-Aware Graph Convolutional Network (GCN)-based method that includes a novel GCN architecture with controlled question-dependent message propagation for the implicit reasoning.
Extensive experiments have been conducted, where QAGCN achieved competitive and even superior performance compared to state-of-the-art explicit-reasoning methods.
Our code and pre-trained models are available in the repository: \url{https://github.com/ruijie-wang-uzh/QAGCN}.

\end{abstract}

\section{Introduction}
\label{section:introduction}

Question answering (QA) over knowledge graphs (KGs)---a task that has been pursued almost since the inception of the Semantic Web (cf. \cite{kaufmann2007})---aims to automatically retrieve answers from KGs for given natural language questions. 
In recent years, multi-relation questions that require reasoning chains over multiple KG triples have been a focus of this task.
An example is ``\textit{who is the mayor of the city where the director of Sleepy Hollow was born},'' which mentions one topic entity (``\textit{Sleepy Hollow}'') and three relations (``\textit{mayor of}'', ``\textit{director of}'', and ``\textit{was born}'').
Correspondingly, starting from the entity \texttt{Sleepy Hollow}, the expected answer \texttt{Jess Talamantes} can be inferred from a reasoning chain that consists of three triples:  [(\texttt{Sleepy Hollow}, \texttt{director}, \texttt{Tim Burton}), (\texttt{Tim Burton}, \texttt{birthplace}, \texttt{Burbank}), (\texttt{Burbank}, \texttt{mayor}, \texttt{Jess Talamantes})]. 
To answer this kind of question, methods with reasoning mechanisms~\cite{DBLP:conf/iclr/DasDZVDKSM18,DBLP:conf/wsdm/QiuWJZ20,DBLP:conf/aaai/ZhangDKSS18,DBLP:conf/coling/ZhouHZ18,DBLP:conf/emnlp/ShiC0LZ21} have been proposed to infer the reasoning chains over KGs step by step.
They typically commence with topic entities in given questions as anchors and try to infer reasoning chains by extending triples according to the semantics of the questions.
This extension is usually performed in two ways: label propagation from entities in the reasoning chain of the current step to other entities that can be added in the next step, and reinforcement learning-based decision-making on choosing entities to add.
Given that the reasoning involves multiple steps and produces explicit states of reasoning chains at each step, we refer to these methods as explicit multi-step reasoning-based.
Recently, they have achieved state-of-the-art (SOTA) performance in the task.
However, their reasoning mechanisms are often complex and difficult to implement or train. 
Furthermore, for better supervision during training, some of these methods~\cite{DBLP:conf/coling/ZhouHZ18} require annotations of reasoning paths, which are usually unavailable in real-world scenarios.

Inspired by the promising performance that graph convolutional networks (GCNs)~\cite{DBLP:journals/kbs/RenLXCWDF22} achieved in learning semantic representations of KG entities, we propose to adapt GCNs for answering multi-relation questions with single-step implicit reasoning that is simpler, more efficient, and easier to adopt than existing reasoning mechanisms.
Specifically, in this paper, we propose a novel Question-Aware GCN-based QA method, called QAGCN, which encodes questions and KG entities in a joint embedding space where questions are close to correct answers (entities). 
The intuition of our method is as follows:
Given a question, if an entity is the correct answer, the reasoning chain of this question would be part of the KG context (i.e., the neighboring triples) of that entity.
For example, considering the above question, the reasoning chain is part of the context of \texttt{Jess Talamantes} within three hops in the KG.
During encoding, we expect the GCN to focus only on messages that pass through the reasoning path and accumulate the messages in the representation of this entity.
We refer to this process as question-aware message propagation.
In this case, the representation of the entity would contain information that is semantically consistent with the given question and can be aligned with the question in the embedding space.
The GCN directly generates semantic representations of the question and KG entities in an end-to-end fashion.
Therefore, we classify it as single-step reasoning.
Also, the reasoning is implicit, given that it is hidden in the message propagation of the GCN.
It is very challenging to achieve this GCN-based reasoning, as the potential KG context of entities could be very large, especially when several hops need to be considered for multi-relation questions.

In summary, we make the following contributions in this paper:
\begin{itemize}
    \item We propose a novel QA method called QAGCN that can answer multi-relation questions via single-step implicit reasoning.
    The method is simpler, more efficient, and easier to adopt than existing reasoning-based methods.
    \item We propose a novel question-aware GCN architecture with contextually controlled message propagation for implicit reasoning.
    \item We conducted extensive experiments to evaluate the effectiveness and efficiency of QAGCN and demonstrate its competitive and even superior performance compared with recent SOTA reasoning-based methods.
\end{itemize}

\section{Related Work}
\label{sec:related_work}

In this section, we first give a detailed introduction to existing reasoning-based methods. Then, we provide a general overview of other QA methods.

As aforementioned, there are primarily two types of reasoning-based methods: label propagation-based and reinforcement learning-based.
Examples of label propagation-based methods include NSM~\cite{DBLP:conf/wsdm/HeL0ZW21}, SR+NSM~\cite{DBLP:conf/acl/ZhangZY000C22}, and TransferNet~\cite{DBLP:conf/emnlp/ShiC0LZ21}.
NSM and SR+NSM label entities in the underlying KG with their probabilities of being part of the reasoning chain of a given question.
In the first step, only topic entities are labeled with positive probabilities.
During reasoning, the probabilities are expected to be passed through the reasoning chain to final answers.
TransferNet also starts with labeling topic entities.
It leverages the semantics of relations in the KG to perform the transfer of labels between entities.
Examples of reinforcement learning-based methods include  MINERVA~\cite{DBLP:conf/iclr/DasDZVDKSM18}, IRN~\cite{DBLP:conf/coling/ZhouHZ18}, and SRN~\cite{DBLP:conf/wsdm/QiuWJZ20}.
MINERVA trains a reinforcement learning agent that walks on KGs from entity to entity through the edges linking them.
Starting from topic entities, the agent is trained to arrive at correct answers after several walking steps.
IRN tries to infer chains of relations that could link topic entities to correct answers based on jointly learned KG embeddings and question encodings. 
SRN employs attentional question encoding and potential-based reward shaping in the training of the reinforcement learning agent.
These methods are the most relevant to our work and will be used as baselines in experiments.

Other important types of QA methods include parsing-based~\cite{DBLP:conf/acl/BerantL14,DBLP:journals/tkde/Hu0YWZ18,kaufmann2007,DBLP:conf/dasfaa/WangWLCCD19,DBLP:journals/corr/abs-2204-08554} and embedding-based~\cite{DBLP:conf/emnlp/BordesCW14,DBLP:journals/corr/BordesUCW15,DBLP:conf/wsdm/HuangZLL19,DBLP:conf/www/LukovnikovFLA17,DBLP:conf/emnlp/SunDZMSC18,DBLP:conf/acl/SaxenaTT20}.
Parsing-based methods construct logical queries, such as SPARQL~\cite{SPARQL} and $\lambda$-DCS~\cite{DBLP:journals/corr/Liang13}, that represent the semantics of given questions and can be evaluated to retrieve answers from KGs.
They can provide parsing results of questions (e.g., dependency tree and intermediate query structures) that can be used to comprehend and assess the QA process of the model. 
However, their performance is usually limited by available resources for question parsing, such as dictionaries, corpora of queries, templates, and heuristics.
Embedding-based methods learn numerical representations of questions and KGs that can be used for direct answer retrieval without logical query construction. 
They can learn the knowledge that parsing-based methods require people to manually define. 
However, most existing methods are limited to answering single-hop questions due to the lack of multi-step reasoning ability.

\section{QAGCN -- A Question-Aware GCN-Based Question Answering Model}
\label{sec:proposed_model}

In this section, we define the multi-relation QA task and elaborate on the proposed QAGCN model.

\subsection{Task Definition}
\label{subsection:task_definition}

A \textbf{knowledge graph (KG)} is denoted as $\mathcal{G} = (E, R, T)$, where $E$ and $R$ denote the entity set and relation set of $\mathcal{G}$, and  $T \subseteq E \times R \times E$ denotes the triple set of $\mathcal{G}$. 
A triple $(e_h, r, e_t)\in T$ denotes that there is a relation $r\in R$ linking the head entity $e_h\in E$ to the tail entity $e_t \in E$. 
Entities and relations are labeled by their names in natural language. 
We denote the set of all labels as $\mathcal{L}$ and define a function $f_l: E \cup R \rightarrow \mathcal{L}$ for retrieving the label of a given entity or relation.

The multi-relation QA task is to train a model that can answer questions from a test set $Q$ over a given KG $\mathcal{G} = (E, R, T)$ based on a disjoint training set $\hat{Q}$. 
Each question $q\in Q \cup \hat{Q}$ is expressed in natural language and can be answered by entities in $E$. 
Also, in line with existing work~\cite{DBLP:conf/wsdm/HeL0ZW21,DBLP:conf/wsdm/QiuWJZ20,DBLP:conf/emnlp/SunBC19,DBLP:conf/coling/ZhouHZ18}, a topic entity is assumed to be annotated.\footnote{If topic entities are not annotated, they can still be easily obtained via named entity recognition, which has been widely studied for decades.} 
The reasoning chain of $q$ is a path $P \subseteq T$ that links the topic entity to the final answers using relations specified in $q$. 

\subsection{The Question-Aware GCN}
\label{sec:question-answer_alignment_module}

\begin{figure}[t]
\includegraphics[width=\textwidth]{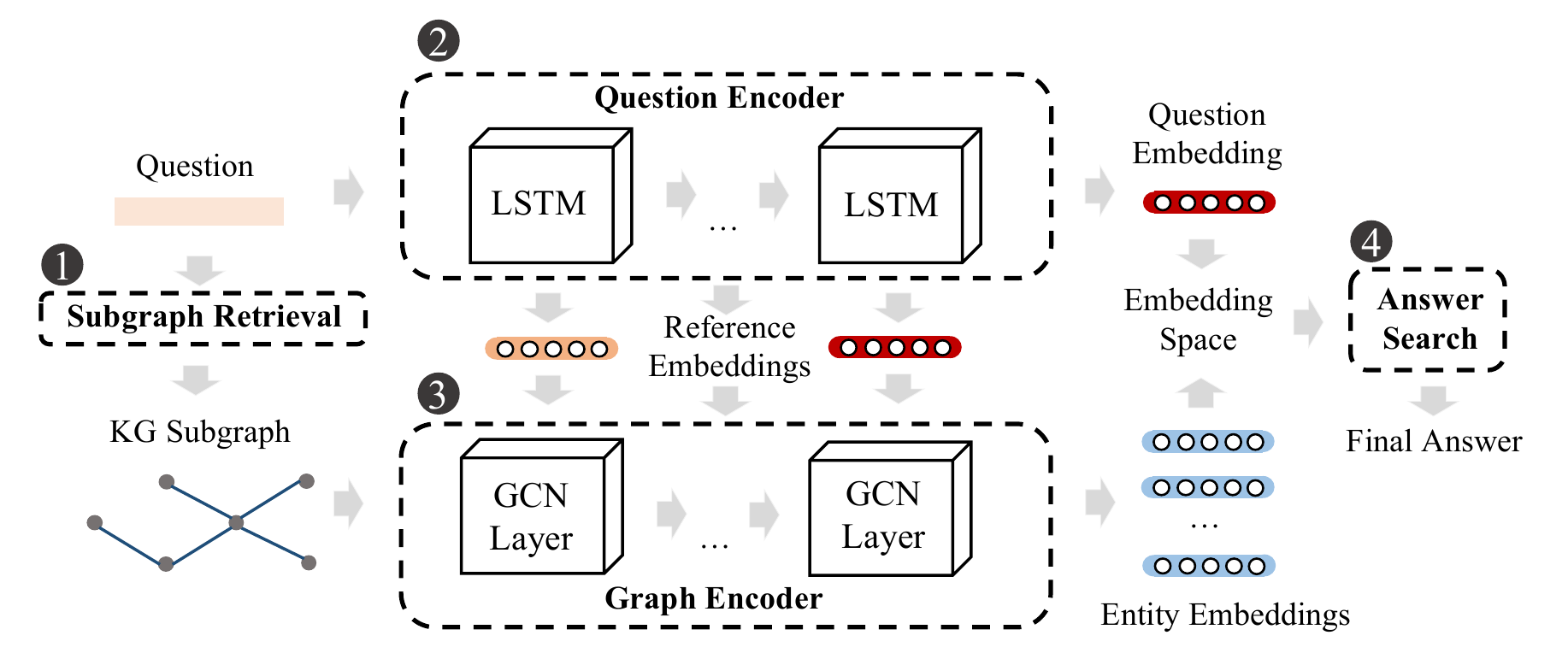}
\caption{An overview of the QAGCN model.} 
\label{fig:alignment_module}
\end{figure}

We present an overview of the proposed question-aware GCN in \cref{fig:alignment_module}.
Given a multi-relation question, we feed it together with a subgraph extracted for the question from the underlying KG as input.
The model consists of a question encoder and a graph encoder that compute semantic representations (embeddings) of the question and subgraph entities, respectively, through several layers of encoding.
We select answers from subgraph entities according to their distances to the question in the output embedding space and further improve the precision of final answers based on relation labels.
The entire process can be segmented into four components: subgraph extraction, question encoding, subgraph encoding, and answer search.
We present detailed elaboration on each of these components in the following.

\subsubsection{Subgraph Extraction} In this component, we aim to extract a more tractable subgraph, which encompasses the reasoning chain of the given question, from the potentially very large underlying KG.
Specifically, given the question $q \in Q \cup \hat{Q}$ posed over the KG $\mathcal{G} = (E, R, T)$, we aim to extract a subgraph $\mathcal{G}_q = (E_q, R_q, T_q)$ that covers the reasoning chain $P$, i.e., $P \subseteq T_q \subseteq T$. 
If $q$ is a $x$-relation question (i.e., $x$ denotes the number of hops of the question), $P$ would consist of $x$ triples linking the topic entity $e_q$ to the answers.
Therefore, we first train a question classifier to predict $x$ for $q$.
The classifier consists of three linear layers with two rectified linear unit (ReLU) layers in between and a softmax output layer. 
For the input of the classifier, we filter out stop words and annotated topic entities from questions. 
Then, the rest words are represented as bag-of-words vectors with two additional entries denoting the number of words and the number of out-of-vocabulary (OOV) words. 
Questions in the training set $\hat{Q}$ are used for vocabulary construction and model training. 
In addition, we remove 15 least used words from the vocabulary to provide training questions containing OOV words.
Based on the estimated $x$ for $q$, in this step, we add all paths of length $x$ or shorter in $\mathcal{G}$ that start from the topic entity $e_q$ to $\mathcal{G}_q$.
Please note that we ignore triple directions to ensure that our method is applicable to different KG schemas. 
For example, the information that Tim Burton directed Sleepy Hollow can be represented as (\texttt{Sleepy Hollow}, \texttt{director}, \texttt{Tim Burton}) or (\texttt{Tim Burton}, \texttt{directed}, \texttt{Sleepy Hollow}), depending on the schema.

\subsubsection{Question Encoding} In this component, we filter out stop words from the question $q$ and segment $q$ into a sequence of $n$ chunks $[c_1, c_2, ..., c_n]$. 
The segmentation is performed by splitting on whitespace, with the exception that the mention of the topic entity $e_q$ remains complete in one chunk.
Then, a pre-trained BERT model~\cite{DBLP:conf/naacl/DevlinCLT19} (bert-base-uncased\footnote{\url{https://huggingface.co/bert-base-uncased}}) is used to encode each chunk into an embedding vector. 
As suggested by bert-as-service\footnote{\url{https://bert-as-service.readthedocs.io/}}, the average pooling of hidden embeddings of the second-to-last BERT layer is computed as the embedding of each chunk. 
We denote the embeddings of $[c_1, c_2, ..., c_n]$ as $[\mathbf{c_1}, \mathbf{c_2}, ..., \mathbf{c_n}]$, where $\mathbf{c_i}\in \mathbb{R}^{d_0}, i = 1, ..., n$, and $d_0$ is the initial embedding size.

These embeddings are fed to $L$ stacked LSTM~\cite{hochreiter1997long} layers, as depicted in \cref{fig:alignment_module}. 
In each layer, there are $n$ LSTM units that sequentially update a cell state passing through them given a sequence of embeddings from the previous layer (or the initial embeddings $[\mathbf{c_1}, \mathbf{c_2}, ..., \mathbf{c_n}]$ in the first layer). 
The computed cell state of the last LSTM layer (denoted as $\mathbf{q_L} \in \mathbb{R}^{d_L}$) is used as the final embedding of $q$, while the cell state of the $l$-th layer, i.e., $\mathbf{q_l} \in \mathbb{R}^{d_l}$, $l = 1, ..., L$, is used as a reference embedding for the computation of attention weights in the $l$-th GCN layer of the graph encoder, which we introduce below.

\subsubsection{Subgraph Encoding} We analogously use a pre-trained BERT model to initialize embeddings of entities and relations in $\mathcal{G}_q$ based on their labels. 
Specifically, each entity $e \in E_q$ and each relation $r \in R_q$ are encoded into $\mathbf{e_0}, \mathbf{r_0} \in \mathbb{R}^{d_0}$ based on their labels $f_l(e)$ and $f_l(r)$, respectively.
Following other GCN work~\cite{DBLP:conf/iclr/KipfW17,DBLP:conf/iclr/VashishthSNT20}, we add inverse edges and self-loops to $\mathcal{G}_q$, i.e., $T_q \leftarrow T_q \cup \{(e_t, r^{-1}, e_h) | (e_h, r, e_t) \in T_q\} \cup \{(e, r^{s}, e) | e\in E_q\}$ and $R_q \leftarrow R_q \cup \{r^{-1} | r \in R_q\} \cup \{r^{s}\}$, where $r^{-1}$ denotes the inverse of $r$, and $r^{s}$ denotes the self-loop relation. 
The inverse relation $r^{-1}$ allows information to be passed from the tail entity $e_t$ to the head entity $e_h$ during graph encoding, and its initial embedding $\mathbf{r^{-1}_0} = -1 \cdot \mathbf{r_0}$.
The self-loop relation $r^{s}$ allows entities to receive information from themselves.
Also, its initial embedding is set to a zero vector, i.e., $\mathbf{r^{s}_0} = \mathbf{0}$. 

The subgraph $\mathcal{G}_q$ with initialized entity and relation embeddings are fed to $L$ stacked GCN layers. And the graph encoding in each layer includes two steps: \emph{message passing} and \emph{message aggregation}. In the \emph{message passing step}, we employ a linear transformation to compute messages that each entity receives from its context. Specifically, for an entity $e$ in the subgraph $\mathcal{G}_q$, its context is defined as a set of all incoming triples $C(e) = \{(\hat{e}, r, e)|(\hat{e}, r, e)\in T_q\}$. In the $l$-th GCN layer, the message that $e$ receives from an incoming triple $(\hat{e}, r, e)\in C(e)$ is computed as follows:
\begin{equation}
\label{equ:message_computation}
    m_l(\hat{e}, r, e) = \mathbf{W_l} \cdot [\mathbf{\hat{e}_{l-1}}\| \mathbf{r_{l-1}}] + \mathbf{b_l},
\end{equation}
where $\mathbf{W_l} \in \mathbb{R}^{d_l \times 2d_{l-1}}$ and $\mathbf{b_l} \in \mathbb{R}^{d_l}$ denote the weight and bias for the message computation, $\mathbf{\hat{e}_{l-1}}, \mathbf{r_{l-1}} \in \mathbb{R}^{d_{l-1}}$ are embeddings of $\hat{e}$ and $r$ computed by the previous GCN layer (or initial embeddings if $l=1$), and $[\mathbf{\hat{e}_{l-1}} \| \mathbf{r_{l-1}}]$ denotes the concatenation of $\mathbf{\hat{e}_{l-1}}$ and $\mathbf{r_{l-1}}$.

Depending on the size of $C(e)$, the entity $e$ may receive a large number of messages in each GCN layer.
However, we expect the model to only consider messages related to the given question, i.e., question-aware message propagation. 
For example, given ``\textit{who is the director of Sleepy Hollow}," only the message from \texttt{Sleepy Hollow} should be considered when encoding \texttt{Tim Burton}. 
To this end, in the \emph{message aggregation step}, we propose an attention mechanism to compute weights for passed messages with reference to the given question. Specifically, in the $l$-th GCN layer, the weight of the message that $e$ receives from $(\hat{e}, r, e)\in C(e)$ is computed as follows:
\begin{equation}
\label{equ:weight_computation}
    w_l(\hat{e}, r, e) = \frac{\exp \bigl( \tanh (\mathbf{A_l} \cdot [m_l(\hat{e}, r, e) \| \mathbf{q_l}])\bigr)}{\sum_{(\hat{e}', r', e) \in C(e)} \exp \bigl( \tanh (\mathbf{A_l} \cdot [m_l(\hat{e}', r', e) \| \mathbf{q_l}]) \bigr)},
\end{equation}
where $\mathbf{q_l} \in \mathbb{R}^{d_l}$ is the reference embedding computed by the $l$-th LSTM layer of the question encoder, $m_l(\cdot)$ computes the message passed to $e$ from an incoming triple according to \cref{equ:message_computation}, and $\mathbf{A_l} \in \mathbb{R}^{1 \times 2 d_l}$ is the weight to learn for the attention mechanism in the $l$-th GCN layer. 
The goal of \cref{equ:weight_computation} is to compute higher weights for messages that are related to the given question while lower weights for irrelevant messages. 

Based on computed weights, the embedding of $e$ in the $l$-th GCN layer is updated as follows:
\begin{equation}
    \mathbf{e_l} = \sum_{(\hat{e}, r, e)\in C(e)} w_l(\hat{e}, r, e) \cdot m_l(\hat{e}, r, e),
\end{equation}
where $w_l(\hat{e}, r, e)\in \mathbb{R}$ and $m_l(\hat{e}, r, e) \in \mathbb{R}^{d_l}$.

\subsubsection{Answer Search} After $L$ GCN layers, each entity $e \in E_q$ is encoded as an embedding $\mathbf{e_L} \in \mathbb{R}^{d_L}$. 
The likelihood of $e$ being the answer of the question $q$ is measured by the Euclidean distance between $\mathbf{e_L}$ and the question embedding $\mathbf{q_L}$. 
We rank all entities in the subgraph $\mathcal{G}_q$ according to computed Euclidean distances to retrieve top-ranked candidate answers, which are denoted as a set $E_c$.
Then, we leverage label information of relations in the KG to filter out outliers in $E_c$ and further improve the precision of our results via the reranking of top-ranked candidate answers.
Specifically, for each entity $e_c \in E_c$, we employ NetworkX and Graph Tool\footnote{\url{https://networkx.org/} and \url{https://graph-tool.skewed.de}} to extract relation paths of the predicted length $x$ that link $e_c$ to the topic entity $e_q$ in the KG subgraph.
The semantics of the relation paths can be assessed based on the labels of constituent relations.
For example, regarding a question estimated to be 2-hop ``\textit{where was the director of Sleepy Hollow born}", we can extract a 2-hop relation path [\texttt{birthplace}, \texttt{director}] for the candidate answer \texttt{Burbank} given triples \{(\texttt{Burbank}, \texttt{birthplace}, \texttt{Tim Burton}), (\texttt{Tim Burton}, \texttt{director}, \texttt{Sleepy Hollow})\}\footnote{We ignore relation directions in this process.} in the subgraph.
Then, the semantics of this path can be examined based on ``\textit{birthplace}'' and ``\textit{director}.''

Specifically, we denote the set of extracted relation paths for each candidate answer $e_c \in E_c$ as $\mathcal{P}_r(e_c)$. 
The plausibility of $e_c$ is evaluated by measuring the semantic similarities between its relation paths and the given question $q$. 
We denote a relation path $P_r \in \mathcal{P}_r(e_c)$ as a sequence of relations, i.e., $P_r = [r_1, ..., r_x]$, where $r_i \in R_q, i = 1, ..., x$. 
Analogous to the initial encoding of question chunks, relations in $P_r$ are also encoded by the pre-trained BERT model, i.e., $P_r$ is encoded as $[\mathbf{r_1}, ..., \mathbf{r_x}]$, where $\mathbf{r_i} \in \mathbb{R}^{d_0}, i = 1, ..., x$.
Then, $[\mathbf{r_1}, ..., \mathbf{r_x}]$ is fed to a single-layer LSTM, and the cell state of the LSTM is used as the embedding of $P_r$, denoted as $\mathbf{P_r} \in \mathbb{R}^{d_p}$, where $d_p$ is the embedding size.
For the given question $q$, we reuse the segmented chunks $[c_1, ..., c_n]$ and their initial embeddings computed in the question encoder, i.e., $[\mathbf{c_1}, ..., \mathbf{c_n}]$, where $\mathbf{c_i}\in \mathbb{R}^{d_0}, i = 1, ..., n$.
The chunk embeddings are fed to another single-layer LSTM, and its cell state is used as the embedding of $q$, i.e., $\mathbf{q_p} \in \mathbb{R}^{d_p}$.
The Euclidean distance between $\mathbf{P_r}$ and $\mathbf{q_p}$ is computed to measure the semantic similarity between $P_r$ and $q$.
We train the above two LSTMs based on training questions in $\hat{Q}$.
Specifically, for each question $\hat{q} \in \hat{Q}$, relation paths between its final answer and topic entity are used as positive samples. 
Using other existing relations in the KG, we randomly generate paths different from the positive ones as negative samples.
The distances between positive samples and $\hat{q}$ are trained to be close to zero, while those for negative samples are trained to be close to one.
For each candidate answer $e_c$, we take the minimum computed distance of all its relation paths in $\mathcal{P}_r(e_c)$ as the final distance of $e_c$. Finally, the candidate answer in $E_c$ with the minimum final distance is selected as the final answer.

\section{Experiments}
\label{sec:experiments}

In this section, we evaluate the effectiveness and efficiency of our model on widely used benchmark datasets, scrutinize the contribution of each component of the model in an ablation study, and present a case study for a better understanding of our model.

\subsection{Effectiveness Evaluation}
\label{subsection:effectiveness_evaluation}

\begin{table}[t]
\centering
\caption{Statistics of adopted benchmark datasets.}
\label{table:data_statistics}
\begin{tabular}{P{2.5cm} P{1.45cm} P{1.45cm} P{1.45cm} P{1.45cm} P{1.45cm} P{1.45cm}}
\toprule
Datasets & \multicolumn{3}{c}{Underlying KG} & \multicolumn{3}{c}{Question Sets} \\
 & \#Entity & \#Relation & \#Triple & \#Train & \#Valid & \#Test \\ \midrule
PQ-2hop & 1,056 & 13 & 1,211 & 1,526 & 190 & 192  \\
PQ-3hop & 1,836 & 13 & 2,839 & 4,158 & 519 & 521  \\
PQL-2hop & 5,034 & 363 & 4,247 & 1,276 & 159 & 159  \\
PQL-3hop & 6,505 & 411 & 5,597 & 825 & 103 & 103  \\
MetaQA 1-hop & 43,234 & 9 & 134,741 & 96,106 & 9,992 & 9,947 \\
MetaQA 2-hop & 43,234 & 9 & 134,741 & 118,980 & 14,872 & 14,872  \\
MetaQA 3-hop & 43,234 & 9 & 134,741 & 114,196 & 14,274 & 14,274  \\
\bottomrule
\end{tabular}
\end{table}

\textbf{Baselines} We first evaluate the overall effectiveness of QAGCN in the multi-relation QA task.
Given that the main goal of this paper is to propose a simple method that is competitive with existing reasoning-based methods that rely on complex reasoning mechanism, we mainly choose reasoning-based QA methods as baselines: MINERVA~\cite{DBLP:conf/iclr/DasDZVDKSM18}, IRN~\cite{DBLP:conf/coling/ZhouHZ18}, SRN~\cite{DBLP:conf/wsdm/QiuWJZ20}, TransferNet~\cite{DBLP:conf/emnlp/ShiC0LZ21}, and NSM~\cite{DBLP:conf/wsdm/HeL0ZW21}.
Among these, MINERVA, IRN, and SRN are reinforcement learning-based.
TransferNet and NSM are label propagation-based.
Also, TransferNet and NSM are state-of-the-art (SOTA) reasoning-based methods in multi-relation QA.
Furthermore, considering that our answer search in the learned embedding space is similar to embedding-based methods, we also select a prominently adopted embedding-based method: EmbedKGQA~\cite{DBLP:conf/acl/SaxenaTT20}.

\noindent \textbf{Datasets} In line with the baselines, we use three collections of datasets that are particularly constructed for the multi-relation QA task: PathQuestion (PQ), PathQuestion-Large (PQL)~\cite{DBLP:conf/coling/ZhouHZ18} and MetaQA~\cite{DBLP:conf/aaai/ZhangDKSS18}.
They consist of QA sets that are named according to the complexity (number of required reasoning hops) of respectively included questions: PQ-2hop, PQ-3hop, PQL-2hop, PQL-3hop, MetaQA 1-hop, MetaQA 2-hop, and MetaQA 3-hop.
PQ and PQL are both open-domain QA datasets based on Freebase~\cite{DBLP:conf/sigmod/BollackerEPST08}, while PQL is a more challenging version of PQ with less training data and larger KGs.
One example question in PQ is: \textit{what is the place of birth of Marguerite Louise Dorleans's other half's kid ?}\footnote{In the original form of this question, all letters are lowercase, the entity phrase is connected by underlines (e.g., marguerite\_louise\_dorleans). We slightly change the format for better readability.} 
MetaQA is constructed based on WikiMovies~\cite{DBLP:conf/emnlp/MillerFDKBW16}, mainly including questions in the movie domain.
One example is: \textit{what genres do the films that share directors with Scarlet Street fall under?}
\cref{table:data_statistics} reports statistics of the QA sets, including the number of included training/validation/test questions and the number of entities/relations/triples in the underlying KG for each question set.

\noindent \textbf{Experimental Details} The experiments on MetaQA and PQ were conducted on a Linux server with two Intel Xeon Gold 6230 CPUs and one NVIDIA GeForce RTX 2080 Ti GPU used by us.
The experiments on PQL were conducted on another Linux server with two AMD EPYC 9124 16-Core CPUs and one NVIDIA GeForce RTX 4090 used by us.
We set hyper-parameters based on grid-search:
Learning rates are set to 0.0002, 0.0005, 0.001, 0.0005, 0.0005, 0.0005, and 0.001 for MetaQA 1,2,3-hop, PQ 2,3-hop, and PQL 2,3-hop, respectively.
Dropouts of 0.1 are used in each of the GCN layers, with the exception of the output layer.

\begin{table}[t]
\centering
\caption{Effectiveness performance on PathQuestion (PQ) and PathQuestion-Large (PQL). (\% Hits@1, best performance in \textbf{bold}, second best \underline{underlined})}
\label{table:effectiveness_1}
\begin{tabular}{P{3.5cm} P{1.7cm} P{1.7cm} P{1.7cm} P{1.7cm}}
\toprule
\multirow{2}{*}{} & PQ-2hop & PQ-3hop & PQL-2hop & PQL-3hop \\ \midrule
MINERVA~\cite{DBLP:conf/iclr/DasDZVDKSM18}  & 75.9  & 71.2 & 71.8 & 65.7 \\ 
IRN~\cite{DBLP:conf/coling/ZhouHZ18} & 91.9  & 83.3  & 63.0 & 61.8 \\ 
EmbedKGQA~\cite{DBLP:conf/acl/SaxenaTT20} & 90.1  & 86.2  & \underline{79.2} & 61.2 \\ 
SRN~\cite{DBLP:conf/wsdm/QiuWJZ20} & \underline{96.3}  & 89.2  & 78.6 & \textbf{77.5} \\
TransferNet~\cite{DBLP:conf/emnlp/ShiC0LZ21}  & 91.1  & \underline{96.5} & 54.7 & 62.1 \\
NSM~\cite{DBLP:conf/wsdm/HeL0ZW21} & 94.2  & \textbf{97.1} & 74.2 & 67.0 \\
\hline
QAGCN & \textbf{98.5}\textsuperscript{*} & 90.6\textsuperscript{*} & \textbf{87.5} & \underline{70.9} \\
\bottomrule
\end{tabular}

Results marked with * are the average results of five runs using different random training/validation/test splits.

\end{table}

\begin{table}[t]
\centering
\caption{Effectiveness performance on MetaQA. (\% Hits@1)}
\label{table:effectiveness_2}
\begin{tabular}{P{3.5cm} P{1.7cm} P{1.7cm} P{1.7cm}}
\toprule
\multirow{2}{*}{} & MetaQA 1-hop & MetaQA 2-hop & MetaQA 3-hop \\ \midrule
MINERVA~\cite{DBLP:conf/iclr/DasDZVDKSM18} & 96.3  & 92.9  & 55.2  \\ 
IRN~\cite{DBLP:conf/coling/ZhouHZ18} & 85.9  & 71.3  & 35.6 \\ 
EmbedKGQA~\cite{DBLP:conf/acl/SaxenaTT20} & \textbf{97.5} & 98.8 & 94.8  \\ 
SRN~\cite{DBLP:conf/wsdm/QiuWJZ20} & 97.0  & 95.1  & 75.2  \\
TransferNet~\cite{DBLP:conf/emnlp/ShiC0LZ21} & \textbf{97.5}  & \textbf{100}  & \textbf{100} \\
NSM~\cite{DBLP:conf/wsdm/HeL0ZW21} & \underline{97.3}  & \underline{99.9}  & \underline{98.9}  \\
\hline
QAGCN & \underline{97.3} & \underline{99.9} & 97.6 \\
\bottomrule
\end{tabular}
\end{table}

\cref{table:effectiveness_1} reports the overall effectiveness performance of our model and other baselines on PQ and PQL, while \cref{table:effectiveness_2} reports the results on MetaQA.
We follow the baselines to measure the performance by Hits@1, which is the percentage of test questions that have been correctly answered by the returned \textit{top-1} answers.
In the above tables, the results of all baselines on MetaQA and the results of MINERVA, IRN, and SRN on PQ and PQL are from existing publications~\cite{DBLP:conf/wsdm/QiuWJZ20,DBLP:conf/wsdm/HeL0ZW21}.
The results of EmbedKGQA, TransferNet, and NSM on PQ and PQL are computed in our experiments with the source code and training configurations released by the original authors.
The following can be observed:

\begin{itemize}
    \item QAGCN achieved superior performance than existing SOTA methods on PQ-2hop and PQL-2hop with relative improvements of 2.3\% and 10.5\%, respectively.
    Also, on MetaQA 1-hop and MetaQA 2-hop, QAGCN achieved the second-best performance, which is very close to the SOTA methods with only relative drops of -0.2\% and -0.1\%.
    This demonstrates the competitiveness of QAGCN, considering its simplicity in comparison with the SOTA methods.

    \item On PQL-3hop, QAGCN is ranked second with a drop of -8.5\% in comparison with SRN.
    This demonstrates the strength of SRN's reinforcement learning(RL)-based policy when answering complex questions.
    However, please note that QAGCN also has a large margin of 5.8\% in comparison with the third-best method NSM.
    This demonstrates that, on complex questions, the simple single-step reasoning of QAGCN could perform better than the SOTA methods with complex multi-step label propagation.

    \item  On PQ-3hop and MetaQA-3hop, QAGCN are both ranked third.
    This reflects that it is indeed challenging for the simple reasoning of QAGCN to answer complex 3-hop questions.
    However, the performance of QAGCN is better than most reasoning-based methods, e.g., 29.8\% and 1.6\% higher than the best-performing RL-based method SRN on MetaQA 3-hop and PQ-3hop, respectively.

\end{itemize}

\subsection{Comparison with SOTA Reasoning-based Method --- NSM}

In this section, we compare with the SOTA reasoning-based method NSM to demonstrate that QAGCN is simpler and easier to train.

\begin{figure}[t]
\includegraphics[width=\textwidth]{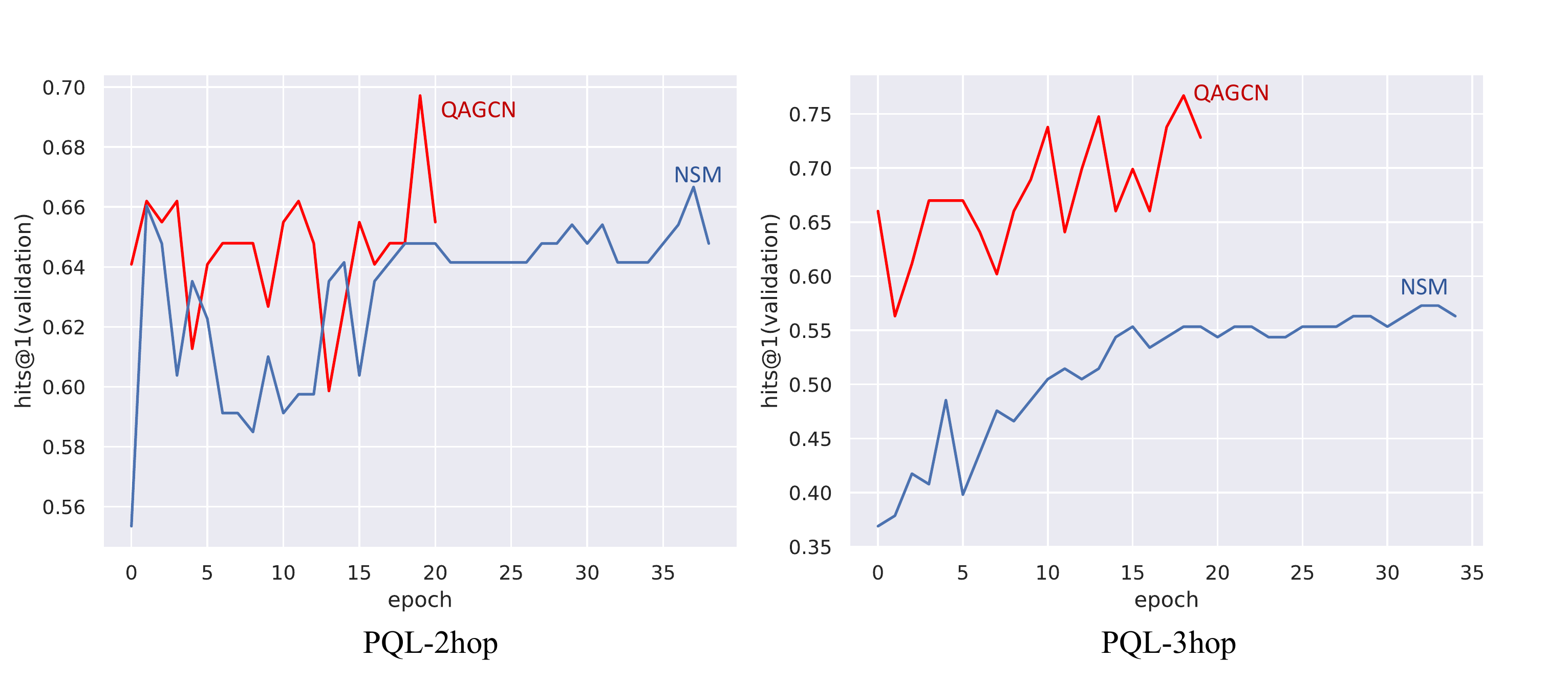}
\caption{Training curves of QAGCN and NSM on PQL-2hop and PQL-3hop.} 
\label{fig:lc}
\end{figure}

The training curves of QAGCN and NSM on PQL-2hop and PQL-3hop are shown in \cref{fig:lc}.
For clarity, we truncate the curves from the second epoch after the optimal epoch of each training.
It can be observed that QAGCN only requires about half of the epochs needed by NSM, specifically 51.4\% and 54.5\% on PQL-2hop and PQL-3hop, respectively.
This demonstrates that, while maintaining the above-demonstrated effectiveness, QAGCN is easier to train than NSM.
The reason is two-fold: First, NSM relies on a teacher network to provide intermediate supervision for the multi-step reasoning of its student network. 
Within the teacher network, two additional multi-step reasoning processes over KGs are performed. 
Therefore, for each training question, NSM needs to reason three times over KGs (once for the student network and twice for the teacher network), which is inefficient given that our model only needs to reason once.
Second, the intermediate supervision of the teacher network is represented as probability distributions of entities. The Kullback-Leibler divergence and Jensen–Shannon divergence~\cite{DBLP:conf/isit/FugledeT04} need to be additionally computed for several pairs of distributions in the loss function of NSM, which makes the model difficult to implement and optimize.

\subsection{Efficiency Evaluation}

\begin{table}[t]
\centering
\caption{Average run-time of each step per test question in ms.
}
\label{table:efficiency}
\begin{tabular}{P{3.5cm} P{1.5cm} P{1.5cm} P{1.5cm} P{1.5cm} P{1.5cm}}
\toprule
\multirow{2}{*}{} & \multicolumn{3}{c}{MetaQA} & \multicolumn{2}{c}{PathQuestion (PQ)} \\ 
\cmidrule(r){2-4}
\cmidrule(l){5-6} 
 & 1-hop & 2-hop & 3-hop & 2-hop & 3-hop \\ \midrule
Question Processing & 1.4 & 1.6 & 1.1 & 0.4 & 0.7 \\
Subgraph Extraction & 3.9 & 6.6 & 357.7 & 1.2 & 18.3 \\
Question Encoding & 16.3 & 16.3 & 10.0 & 8.4 & 9.9 \\
Subgraph Encoding & 1.1 & 2.9 & 26.5 & 1.3 & 2.4 \\
Answer Search & 0.6 & 5.2 & 287.1 & 0.4 & 2.3 \\
Answer Re-ranking & 79.5 & 123.2 & 1012.1 & 8.0 & 15.4 \\
Total & 102.8 & 155.8 & 1694.5 & 19.7 & 49.0 \\
\bottomrule
\end{tabular}
\end{table}

Question answering is supposed to be an online service that answers questions in real-time. 
We evaluated the average time cost of each step of our model on MetaQA and PathQuestion.
The results are reported in \cref{table:efficiency}. 
In this table, Question Processing includes the segmentation and classification of questions. 
The others are in accordance with the steps introduced in \cref{sec:proposed_model}.
Please note that we list the path extraction-based answer refinement separately as Answer Re-ranking, given that the explicit path extraction in this step could be particularly time-consuming and is worth specific attention.
Most time costs are below or near 100ms, which can be seen as instantaneous in user interfaces~\cite{Nielsen91}. Only 3-hop questions in MetaQA require an average of 1.7 seconds, which we consider acceptable when contrasted with the complexity of 3-hop questions.
It is worth mentioning that our implementation can be 
further optimized for a real-world deployment.
For example, the subgraph extraction can be performed offline given a KG, and the extraction and encoding of relation paths for candidate answers can be performed in a completely parallel way.

\subsection{Ablation Study}

\begin{table}[t]
\centering
\caption{Model performance changes in the ablation study. (\% Hits@1)}
\label{table:ablation}
\begin{tabular}{P{3.6cm} P{1.5cm} P{1.5cm} P{1.5cm} P{1.5cm} P{1.5cm}}
\toprule
\multirow{2}{*}{} & \multicolumn{3}{c}{MetaQA} & \multicolumn{2}{c}{PathQuestion (PQ)} \\ 
\cmidrule(r){2-4}
\cmidrule(l){5-6}
 & 1-hop & 2-hop & 3-hop & 2-hop & 3-hop \\ 
 \midrule
Complete Model & 97.3 & 99.9 & 97.6 & 98.4\textsuperscript{*} & 92.1\textsuperscript{*} \\
Re-ranking Removed & 95.8 & 87.4 & 55.3 & 72.9 & 51.4 \\
Q\&G-Encoders Removed & 97.0 & 91.3 & 86.9 & 74.5 & 43.2 \\ 
\midrule
G-Encoder (Conv) & 64.6 & 63.9 & 40.8 & 67.7 & 28.0 \\
G-Encoder (Linear) & 72.0 & 55.6 & 15.2 & 71.4 & 43.0 \\
Q\&G-Encoders (Linear) & 70.7 & 52.4 & 0.1 & 63.5 & 16.9 \\
\bottomrule
\end{tabular}

* We report the results of QAGCN on the first run of PQ-2/3hop.
Hence, these results differ from the averages reported in \cref{table:effectiveness_1}.

\end{table}

To examine the contribution of each component of our model, we evaluate the model in a series of degraded scenarios and report the performance changes in \cref{table:ablation}. 
\textbf{Complete Model} reports the results of the complete QAGCN model. \textbf{Re-ranking Removed} shows the performance when candidate answers are not re-ranked in Answer Search. 
The importance of the re-ranking is demonstrated by the performance degradation (e.g., -43.3\% on MetaQA 3-hop and -44.2\% on PathQuestion 3-hop). 
In \textbf{Q\&G-Encoders Removed}, the question and graph encoders are removed, and we use the pre-trained BERT model to directly encode questions and entities.
Top-\textit{k} candidate answers are retrieved and then re-ranked by the original re-ranking module.\footnote{The values of \textit{k} were set to be consistent with those used in the complete model.} 
In this scenario, there is also a significant performance drop (e.g., -11.0\% on MetaQA 3-hop and -53.1\% on PathQuestion 3-hop). 
From the above three scenarios, we observe that both the re-ranking module and question/graph encoders are necessary, and that they are complementary to each other. 

In \textbf{G-Encoder (Conv)}, we replace the proposed attentional GCNs with conventional GCNs~\cite{DBLP:conf/iclr/KipfW17}. 
The question encoder remains unchanged, and we report direct results without re-ranking. 
Compared to Re-ranking Removed, i.e., the original encoders, performance degradations can be observed on all datasets (e.g., -26.2\% on MetaQA 3-hop and -45.5\% on PathQuestion 3-hop). 
In \textbf{G-Encoder (Linear)}, we further replace the conventional GCNs with feedforward neural networks. 
In this scenario, entities are encoded without considering their contexts in KGs. 
Compared with G-Encoder (Conv), performance drops can be observed on questions with large subgraphs (e.g., -13.0\% on MetaQA 2-hop and -62.7\% on MetaQA 3-hop). 
While for questions with small subgraphs, in which entities do not have sufficient contexts, improvements are observed (e.g., +11.5\% on MetaQA 1-hop and +5.5\% on PathQuestion 2-hop).
In \textbf{Q\&G-Encoders (Linear)}, based on G-Encoder (Linear), we further replace the LSTMs in the question encoder with feedforward neural networks. 
Further performance degradations can be observed on all datasets (e.g., -99.3\% on MetaQA 3-hop and -60.7\% on PathQuestion 3-hop).

\subsection{Case Study with Embedding Visualization}
\label{sec:case_study}
For an intuitive understanding of the proposed QAGCN model, we select an example question to conduct a case study with the visualization of question and entity embeddings. 
Specifically, we picked 
the following 3-hop test question from PathQuestion: \textit{Isabella of Portugal's child's spouse's place of death?} The topic entity is \texttt{Isabella of Portugal} with a support path of three triples:
\\ \centerline{\{ (\texttt{Isabella of Portugal}, \texttt{children}, \texttt{Maria of Spain}),}
\\  \centerline{(\texttt{Maria of Spain}, \texttt{spouse}, \texttt{Maximilian II Holy Roman Emperor}),}
\\ \centerline{(\texttt{Maximilian II Holy Roman Emperor}, \texttt{place of death}, \texttt{Regensburg}) \}.}

For this question, three GCN layers are used in the graph encoder to encode entities in the subgraph that covers all paths of length one, two, and three starting from the topic entity \texttt{Isabella of Portugal}. 
The question is accordingly encoded by the question encoder with three layers of LSTMs. 
Then, according to Euclidean distances between the question embedding and entity embeddings, we obtain the top-3 candidate answers: \texttt{Vienna} (distance: 1.53), \texttt{Regensburg} (distance: 1.89), and \texttt{Catholicism} (distance: 3.17).
The correct answer \texttt{Regensburg} is ranked after \texttt{Vienna}. 
Since \texttt{Vienna} is the place of birth of \texttt{Maximilian II Holy Roman Emperor}, it is likely that this trained model struggles to differentiate between ``birth" and ``death." 
A second possible reason is that this trained model is biased towards \texttt{Vienna} given that \texttt{Vienna} is the answer of 12 training questions that ask for the place of death of people, while \texttt{Regensburg} is the answer of only one training question.

\begin{figure}[t]
\includegraphics[width=\textwidth]{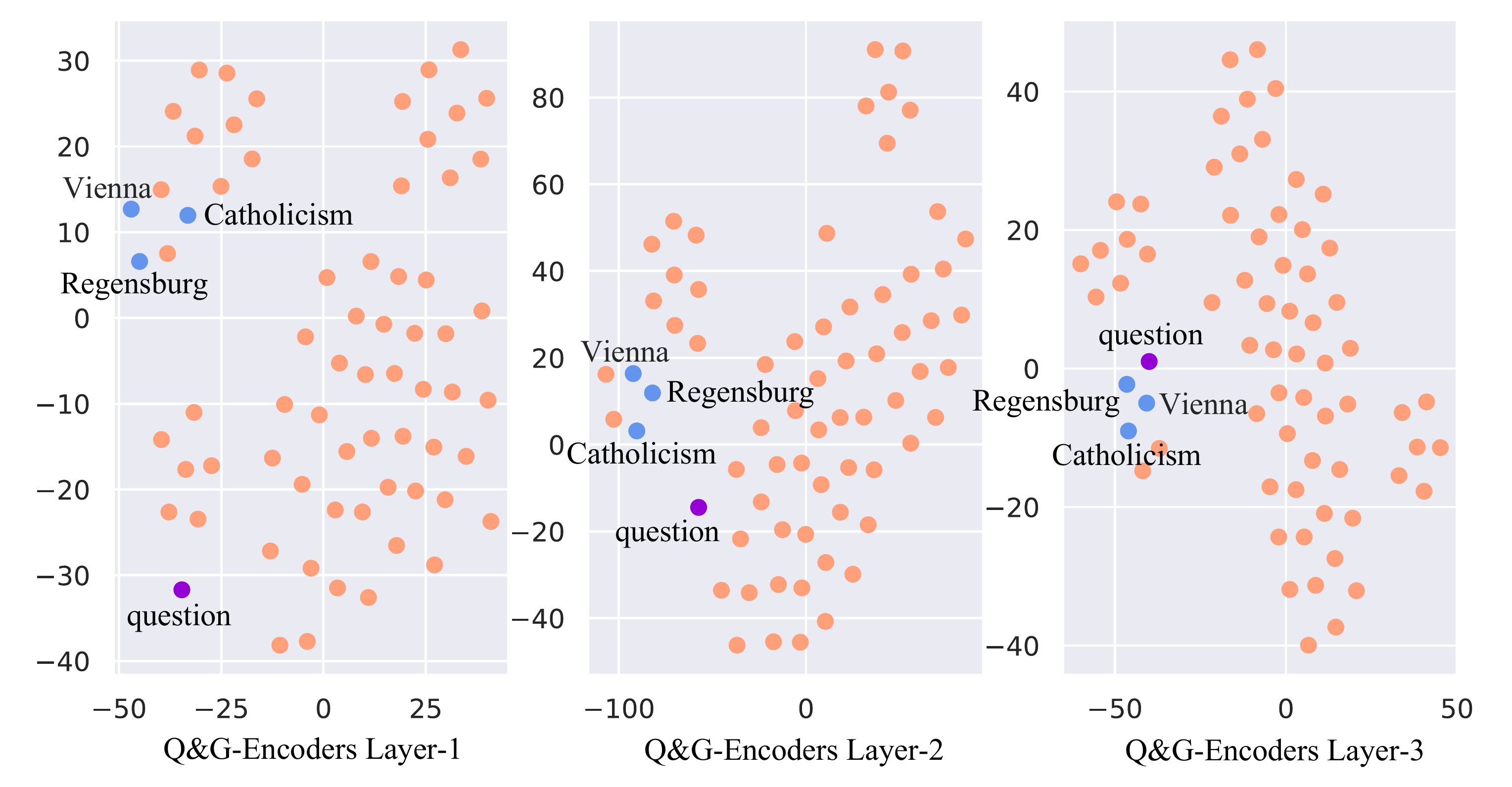}
\caption{Visualization of the embedding space after each layer of the Q\&G-Encoders. Please note that the axes do not have a same scale.
The  \colorsquare{9400D3}{purple}{0.8em}{0.8em} point indicates the question embedding after each layer.
The \colorsquare{6495ED}{blue}{0.8em}{0.8em} points indicate the top-3 candidates finally retrieved.
The \colorsquare{FFA07A}{orange}{0.8em}{0.8em} points indicate all other entities in the subgraph.
}
\label{fig:embedding_visualization}
\end{figure}

To visualize the joint embedding space of the LSTM-based question encoder and GCN-based graph encoder for each of the three layers, we use t-SNE~\cite{JMLR:v9:vandermaaten08a} to compute dimension-reduced entity and question embeddings and plot them in \cref{fig:embedding_visualization}.
We highlight the final three top candidate answers in blue and plot all other entities in the subgraph as orange points. 
In addition, we plot the location of the question embedding after each of the LSTM layers with a purple point. 
It can be observed that initially (i.e., in Layer-1) the correct answer is not at all close to the question in the embedding space. 
However, after more layers of encoding, more information consistent with the given question is passed to \texttt{Regensburg} by the graph encoder, and the question embedding is accordingly transformed by the question encoder. 
Therefore, we can observe that \texttt{Regensburg} and the question approach each other gradually in the embedding space.

Then, we extract relation paths between the top-3 candidate answers and the topic entity, which are respectively \{[\texttt{place of birth}, \texttt{spouse}, \texttt{children}]\}, \{[\texttt{place of death}, \texttt{spouse}, \texttt{children}]\}, and \{[\texttt{religion}, \texttt{spouse}, \texttt{children}], [\texttt{religion}, \texttt{parents}, \texttt{children}]\} for \texttt{Vienna}, \texttt{Regensburg}, and \texttt{Catholicism}. 
The minimum path-question distances of \texttt{Vienna}, \texttt{Regensburg}, and \texttt{Catholicism} computed by the re-ranking module are respectively 0.99, 0.69, and 1.00. Finally, with the minimum distance 0.69, \texttt{Regensburg} is correctly returned as the final answer.

\section{Conclusion and outlook}
\label{sec:conclusion_and_outlook}

In this paper, we propose a novel QA model called QAGCN, which answers multi-relation questions via single-step implicit reasoning over KGs.
A novel question-aware GCN is proposed to perform question-dependent message propagation while encoding KGs.
It is able to represent questions and KG entities in a joint embedding space, where questions are close to their answers.
Our model has been demonstrated to be competitive and even superior against SOTA reasoning-based methods in experiments, while our model is simpler and easier to train and implement. 
The efficiency and contribution of each component of our model have also been examined and analyzed in detail.

The task we solve in this paper is in line with other existing QA work that assumes all given questions are answerable.
Therefore, if the answer to a question does not exist in the accompanied KG, our model will still confidently report an answer, which is actually incorrect. 
In future work, we plan to investigate a method that can detect if a given question is answerable regarding a given KG.

\section*{Acknowledgements}
This work has been partially supported by the University Research Priority Program ``Dynamics of Healthy Aging'' at the University of Zurich and the Swiss National Science Foundation through project MediaGraph (contract no.\ 202125).
Michael Cochez is partially funded by the Graph-Massivizer project, funded by the Horizon Europe programme of the European Union (grant 101093202).

%
%
%
\bibliographystyle{splncs04}
\bibliography{bibliography}

\begin{thebibliography}{10}
\providecommand{\url}[1]{\texttt{#1}}
\providecommand{\urlprefix}{URL }
\providecommand{\doi}[1]{https://doi.org/#1}

\bibitem{DBLP:conf/acl/BerantL14}
Berant, J., Liang, P.: Semantic parsing via paraphrasing. In: Proceedings of the 52nd Annual Meeting of the Association for Computational Linguistics, {ACL} 2014, June 22-27, 2014, Baltimore, MD, USA, Volume 1: Long Papers. pp. 1415--1425. The Association for Computer Linguistics (2014). \doi{10.3115/v1/p14-1133}

\bibitem{DBLP:conf/sigmod/BollackerEPST08}
Bollacker, K.D., Evans, C., Paritosh, P.K., Sturge, T., Taylor, J.: Freebase: a collaboratively created graph database for structuring human knowledge. In: Wang, J.T. (ed.) Proceedings of the {ACM} {SIGMOD} International Conference on Management of Data, {SIGMOD} 2008, Vancouver, BC, Canada, June 10-12, 2008. pp. 1247--1250. {ACM} (2008). \doi{10.1145/1376616.1376746}

\bibitem{DBLP:conf/emnlp/BordesCW14}
Bordes, A., Chopra, S., Weston, J.: Question answering with subgraph embeddings. In: Moschitti, A., Pang, B., Daelemans, W. (eds.) Proceedings of the 2014 Conference on Empirical Methods in Natural Language Processing, {EMNLP} 2014, October 25-29, 2014, Doha, Qatar, {A} meeting of SIGDAT, a Special Interest Group of the {ACL}. pp. 615--620. {ACL} (2014). \doi{10.3115/v1/d14-1067}

\bibitem{DBLP:journals/corr/BordesUCW15}
Bordes, A., Usunier, N., Chopra, S., Weston, J.: Large-scale simple question answering with memory networks. CoRR  \textbf{abs/1506.02075} (2015), \url{http://arxiv.org/abs/1506.02075}

\bibitem{DBLP:conf/iclr/DasDZVDKSM18}
Das, R., Dhuliawala, S., Zaheer, M., Vilnis, L., Durugkar, I., Krishnamurthy, A., Smola, A., McCallum, A.: Go for a walk and arrive at the answer: Reasoning over paths in knowledge bases using reinforcement learning. In: 6th International Conference on Learning Representations, {ICLR} 2018, Vancouver, BC, Canada, April 30 - May 3, 2018, Conference Track Proceedings. OpenReview.net (2018), \url{https://openreview.net/forum?id=Syg-YfWCW}

\bibitem{DBLP:conf/naacl/DevlinCLT19}
Devlin, J., Chang, M., Lee, K., Toutanova, K.: {BERT:} pre-training of deep bidirectional transformers for language understanding. In: Burstein, J., Doran, C., Solorio, T. (eds.) Proceedings of the 2019 Conference of the North American Chapter of the Association for Computational Linguistics: Human Language Technologies, {NAACL-HLT} 2019, Minneapolis, MN, USA, June 2-7, 2019, Volume 1 (Long and Short Papers). pp. 4171--4186. Association for Computational Linguistics (2019). \doi{10.18653/v1/n19-1423}

\bibitem{DBLP:conf/isit/FugledeT04}
Fuglede, B., Tops{\o}e, F.: Jensen-shannon divergence and hilbert space embedding. In: Proceedings of the 2004 {IEEE} International Symposium on Information Theory, {ISIT} 2004, Chicago Downtown Marriott, Chicago, Illinois, USA, June 27 - July 2, 2004. p.~31. {IEEE} (2004). \doi{10.1109/ISIT.2004.1365067}

\bibitem{SPARQL}
Harris, S., Seaborne, A., Prud'hommeaux, E.: Sparql 1.1 query language. w3c recommendation (2013), \url{https://www.w3.org/TR/sparql11-query/}

\bibitem{DBLP:conf/wsdm/HeL0ZW21}
He, G., Lan, Y., Jiang, J., Zhao, W.X., Wen, J.: Improving multi-hop knowledge base question answering by learning intermediate supervision signals. In: Lewin{-}Eytan, L., Carmel, D., Yom{-}Tov, E., Agichtein, E., Gabrilovich, E. (eds.) {WSDM} '21, The Fourteenth {ACM} International Conference on Web Search and Data Mining, Virtual Event, Israel, March 8-12, 2021. pp. 553--561. {ACM} (2021). \doi{10.1145/3437963.3441753}

\bibitem{hochreiter1997long}
Hochreiter, S., Schmidhuber, J.: Long short-term memory. Neural computation  \textbf{9}(8),  1735--1780 (1997)

\bibitem{DBLP:journals/tkde/Hu0YWZ18}
Hu, S., Zou, L., Yu, J.X., Wang, H., Zhao, D.: Answering natural language questions by subgraph matching over knowledge graphs. {IEEE} Trans. Knowl. Data Eng.  \textbf{30}(5),  824--837 (2018). \doi{10.1109/TKDE.2017.2766634}

\bibitem{DBLP:conf/wsdm/HuangZLL19}
Huang, X., Zhang, J., Li, D., Li, P.: Knowledge graph embedding based question answering. In: Culpepper, J.S., Moffat, A., Bennett, P.N., Lerman, K. (eds.) Proceedings of the Twelfth {ACM} International Conference on Web Search and Data Mining, {WSDM} 2019, Melbourne, VIC, Australia, February 11-15, 2019. pp. 105--113. {ACM} (2019). \doi{10.1145/3289600.3290956}

\bibitem{kaufmann2007}
Kaufmann, E., Bernstein, A.: How useful are natural language interfaces to the semantic web for casual end-users? In: Aberer, K., Choi, K.S., Noy, N., Allemang, D., Lee, K.I., Nixon, L., Golbeck, J., Mika, P., Maynard, D., Mizoguchi, R., Schreiber, G., Cudr{\'e}-Mauroux, P. (eds.) The Semantic Web. pp. 281--294. Springer Berlin Heidelberg, Berlin, Heidelberg (2007)

\bibitem{DBLP:conf/iclr/KipfW17}
Kipf, T.N., Welling, M.: Semi-supervised classification with graph convolutional networks. In: 5th International Conference on Learning Representations, {ICLR} 2017, Toulon, France, April 24-26, 2017, Conference Track Proceedings. OpenReview.net (2017), \url{https://openreview.net/forum?id=SJU4ayYgl}

\bibitem{DBLP:journals/corr/Liang13}
Liang, P.: Lambda dependency-based compositional semantics. CoRR  \textbf{abs/1309.4408} (2013), \url{http://arxiv.org/abs/1309.4408}

\bibitem{DBLP:conf/www/LukovnikovFLA17}
Lukovnikov, D., Fischer, A., Lehmann, J., Auer, S.: Neural network-based question answering over knowledge graphs on word and character level. In: Barrett, R., Cummings, R., Agichtein, E., Gabrilovich, E. (eds.) Proceedings of the 26th International Conference on World Wide Web, {WWW} 2017, Perth, Australia, April 3-7, 2017. pp. 1211--1220. {ACM} (2017). \doi{10.1145/3038912.3052675}

\bibitem{JMLR:v9:vandermaaten08a}
van~der Maaten, L., Hinton, G.: Visualizing data using t-sne. Journal of Machine Learning Research  \textbf{9}(86),  2579--2605 (2008), \url{http://jmlr.org/papers/v9/vandermaaten08a.html}

\bibitem{DBLP:conf/emnlp/MillerFDKBW16}
Miller, A.H., Fisch, A., Dodge, J., Karimi, A., Bordes, A., Weston, J.: Key-value memory networks for directly reading documents. In: Su, J., Carreras, X., Duh, K. (eds.) Proceedings of the 2016 Conference on Empirical Methods in Natural Language Processing, {EMNLP} 2016, Austin, Texas, USA, November 1-4, 2016. pp. 1400--1409. The Association for Computational Linguistics (2016). \doi{10.18653/v1/d16-1147}

\bibitem{Nielsen91}
Nielsen, J.: Response times: The 3 important limits (1991), \url{https://www.nngroup.com/articles/response-times-3-important-limits/}

\bibitem{DBLP:conf/wsdm/QiuWJZ20}
Qiu, Y., Wang, Y., Jin, X., Zhang, K.: Stepwise reasoning for multi-relation question answering over knowledge graph with weak supervision. In: Caverlee, J., Hu, X.B., Lalmas, M., Wang, W. (eds.) {WSDM} '20: The Thirteenth {ACM} International Conference on Web Search and Data Mining, Houston, TX, USA, February 3-7, 2020. pp. 474--482. {ACM} (2020). \doi{10.1145/3336191.3371812}

\bibitem{DBLP:journals/kbs/RenLXCWDF22}
Ren, H., Lu, W., Xiao, Y., Chang, X., Wang, X., Dong, Z., Fang, D.: Graph convolutional networks in language and vision: {A} survey. Knowl. Based Syst.  \textbf{251},  109250 (2022). \doi{10.1016/J.KNOSYS.2022.109250}, \url{https://doi.org/10.1016/j.knosys.2022.109250}

\bibitem{DBLP:conf/acl/SaxenaTT20}
Saxena, A., Tripathi, A., Talukdar, P.P.: Improving multi-hop question answering over knowledge graphs using knowledge base embeddings. In: Proceedings of the 58th Annual Meeting of the Association for Computational Linguistics, {ACL} 2020, Online, July 5-10, 2020. pp. 4498--4507. Association for Computational Linguistics (2020). \doi{10.18653/v1/2020.acl-main.412}, \url{https://doi.org/10.18653/v1/2020.acl-main.412}

\bibitem{DBLP:conf/emnlp/ShiC0LZ21}
Shi, J., Cao, S., Hou, L., Li, J., Zhang, H.: Transfernet: An effective and transparent framework for multi-hop question answering over relation graph. In: Proceedings of the 2021 Conference on Empirical Methods in Natural Language Processing, {EMNLP} 2021, Virtual Event / Punta Cana, Dominican Republic, 7-11 November, 2021. pp. 4149--4158. Association for Computational Linguistics (2021). \doi{10.18653/v1/2021.emnlp-main.341}, \url{https://doi.org/10.18653/v1/2021.emnlp-main.341}

\bibitem{DBLP:conf/emnlp/SunBC19}
Sun, H., Bedrax{-}Weiss, T., Cohen, W.W.: Pullnet: Open domain question answering with iterative retrieval on knowledge bases and text. In: Inui, K., Jiang, J., Ng, V., Wan, X. (eds.) Proceedings of the 2019 Conference on Empirical Methods in Natural Language Processing and the 9th International Joint Conference on Natural Language Processing, {EMNLP-IJCNLP} 2019, Hong Kong, China, November 3-7, 2019. pp. 2380--2390. Association for Computational Linguistics (2019). \doi{10.18653/v1/D19-1242}

\bibitem{DBLP:conf/emnlp/SunDZMSC18}
Sun, H., Dhingra, B., Zaheer, M., Mazaitis, K., Salakhutdinov, R., Cohen, W.W.: Open domain question answering using early fusion of knowledge bases and text. In: Riloff, E., Chiang, D., Hockenmaier, J., Tsujii, J. (eds.) Proceedings of the 2018 Conference on Empirical Methods in Natural Language Processing, Brussels, Belgium, October 31 - November 4, 2018. pp. 4231--4242. Association for Computational Linguistics (2018). \doi{10.18653/v1/d18-1455}

\bibitem{DBLP:journals/corr/abs-2204-08554}
Thai, D., Ravishankar, S., Abdelaziz, I., Chaudhary, M., Mihindukulasooriya, N., Naseem, T., Das, R., Kapanipathi, P., Fokoue, A., McCallum, A.: Cbr-ikb: {A} case-based reasoning approach for question answering over incomplete knowledge bases. CoRR  \textbf{abs/2204.08554} (2022). \doi{10.48550/ARXIV.2204.08554}, \url{https://doi.org/10.48550/arXiv.2204.08554}

\bibitem{DBLP:conf/iclr/VashishthSNT20}
Vashishth, S., Sanyal, S., Nitin, V., Talukdar, P.P.: Composition-based multi-relational graph convolutional networks. In: 8th International Conference on Learning Representations, {ICLR} 2020, Addis Ababa, Ethiopia, April 26-30, 2020. OpenReview.net (2020), \url{https://openreview.net/forum?id=BylA\_C4tPr}

\bibitem{DBLP:conf/dasfaa/WangWLCCD19}
Wang, R., Wang, M., Liu, J., Chen, W., Cochez, M., Decker, S.: Leveraging knowledge graph embeddings for natural language question answering. In: Li, G., Yang, J., Gama, J., Natwichai, J., Tong, Y. (eds.) Database Systems for Advanced Applications - 24th International Conference, {DASFAA} 2019, Chiang Mai, Thailand, April 22-25, 2019, Proceedings, Part {I}. Lecture Notes in Computer Science, vol. 11446, pp. 659--675. Springer (2019). \doi{10.1007/978-3-030-18576-3\_39}

\bibitem{DBLP:conf/acl/ZhangZY000C22}
Zhang, J., Zhang, X., Yu, J., Tang, J., Tang, J., Li, C., Chen, H.: Subgraph retrieval enhanced model for multi-hop knowledge base question answering. In: Muresan, S., Nakov, P., Villavicencio, A. (eds.) Proceedings of the 60th Annual Meeting of the Association for Computational Linguistics (Volume 1: Long Papers), {ACL} 2022, Dublin, Ireland, May 22-27, 2022. pp. 5773--5784. Association for Computational Linguistics (2022). \doi{10.18653/V1/2022.ACL-LONG.396}, \url{https://doi.org/10.18653/v1/2022.acl-long.396}

\bibitem{DBLP:conf/aaai/ZhangDKSS18}
Zhang, Y., Dai, H., Kozareva, Z., Smola, A.J., Song, L.: Variational reasoning for question answering with knowledge graph. In: McIlraith, S.A., Weinberger, K.Q. (eds.) Proceedings of the Thirty-Second {AAAI} Conference on Artificial Intelligence, (AAAI-18), the 30th innovative Applications of Artificial Intelligence (IAAI-18), and the 8th {AAAI} Symposium on Educational Advances in Artificial Intelligence (EAAI-18), New Orleans, Louisiana, USA, February 2-7, 2018. pp. 6069--6076. {AAAI} Press (2018), \url{https://www.aaai.org/ocs/index.php/AAAI/AAAI18/paper/view/16983}

\bibitem{DBLP:conf/coling/ZhouHZ18}
Zhou, M., Huang, M., Zhu, X.: An interpretable reasoning network for multi-relation question answering. In: Bender, E.M., Derczynski, L., Isabelle, P. (eds.) Proceedings of the 27th International Conference on Computational Linguistics, {COLING} 2018, Santa Fe, New Mexico, USA, August 20-26, 2018. pp. 2010--2022. Association for Computational Linguistics (2018), \url{https://aclanthology.org/C18-1171/}

\end{thebibliography}

\end{document}